\pdfoutput=1

\documentclass[11pt]{article}

\usepackage[]{acl}

\usepackage{times}
\usepackage{latexsym}
\usepackage{amsfonts,amssymb}
\usepackage{multirow}
\usepackage[T1]{fontenc}
\usepackage{booktabs}
\usepackage{amsmath}
\usepackage{hyperref}
\usepackage{MnSymbol}
\usepackage[utf8]{inputenc}
\usepackage{subfigure}
\usepackage{graphicx}
\usepackage{float}
\usepackage{caption}
\usepackage{diagbox}
\usepackage{makecell}
\usepackage{bbding} 
\usepackage{microtype}

\usepackage{inconsolata}
\usepackage{pifont}

\newcommand{\lei}[1]{\textcolor{orange}{[#1 - lei]}}
\newcommand{\lyx}[1]{\textcolor{green}{[#1 - lyx]}}
\title{Towards Multimodal Video Paragraph Captioning Models \\Robust to Missing Modality}

\newcommand*{\affaddr}[1]{#1}
\newcommand*{\affmark}[1][*]{\textsuperscript{#1}}
\author{
  Sishuo Chen\affmark[\P], 
  Lei Li\affmark[\dag],
  Shuhuai Ren\affmark[\S],
  Rundong Gao\affmark[\P],\\
  \textbf{Yuanxin Liu\affmark[\S]},
  \textbf{Xiaohan Bi\affmark[\P],
  Xu Sun\affmark[\S],
  Lu Hou\affmark[\ddag]} \\
\affaddr{\affmark[\P] Center for Data Science, Peking University} \\
  \affaddr{\affmark[\S] National Key Laboratory for Multimedia Information Processing,\\
      School of Computer Science, Peking University}\\
  \affaddr{\affmark[\dag] The University of Hong Kong} \quad \affaddr{\affmark[\ddag] Huawei Noah's Ark Lab} \\
  \texttt{\small\{chensishuo, xusun\}@pku.edu.cn} \quad 
  \texttt{\small{nlp.lilei@gmail.com}}\\ 
  \texttt{\small\{shuhuai\_ren, gaord20, liuyuanxin, bxh\}@stu.pku.edu.cn} \quad 
  \texttt{\small{houlu3@huawei.com}} \\  
}

\begin{document}
\maketitle

\vspace*{0.5cm}

\begin{abstract}
Video paragraph captioning (VPC) involves generating detailed narratives for long videos, utilizing supportive modalities such as speech and event boundaries. 
However, the existing models are constrained by the assumption of constant availability of a single auxiliary modality, which is impractical given the diversity and unpredictable nature of real-world scenarios.
To this end, we propose a Missing-Resistant framework MR-VPC that effectively harnesses all available auxiliary inputs and maintains resilience even in the absence of certain modalities. 
Under this framework, we propose the Multimodal VPC (MVPC) architecture integrating video, speech, and event boundary inputs in a unified manner to process various auxiliary inputs.
Moreover, to fortify the model against incomplete data, we introduce \emph{DropAM}, a data augmentation strategy that randomly omits auxiliary inputs, paired with \emph{DistillAM}, a regularization target that distills knowledge from teacher models trained on modality-complete data, enabling efficient learning in modality-deficient environments.
Through exhaustive experimentation on YouCook2 and ActivityNet Captions, MR-VPC has proven to deliver superior performance on modality-complete and modality-missing test data.
This work highlights the significance of developing resilient VPC models and paves the way for more adaptive, robust multimodal video understanding.\footnote{Our code is available at \url{https://github.com/lancopku/MR-VPC}.}

\end{abstract}


\section{Introduction}
Video Paragraph Captioning (VPC)~\citep{park2019adversarial} is a fundamental video-language understanding task that requires the model to generate paragraph-level captions for minutes-long videos.
Besides raw video frames, there exist several auxiliary modalities that can potentially serve as supplementary inputs, such as speech inputs utilized in Vid2Seq~\citep{yang2023vid2seq}, flow features used in MART~\citep{lei2020mart}, and event boundaries (the start and end timestamps of the events) leveraged in various models~\citep[etc]{zhou2018end,yamazaki2022vlcap,yamazaki2022vltint}.
Despite the growing performance of these models, we notice that they assume to have access to the same auxiliary modality during both training and testing, which contradicts reality.
In real-world scenarios, the availability of modalities undergoes dynamic changes, which leads to the following two issues for the models developed under the unrealistic assumption.\looseness=-1

\begin{figure}[t]
\centering
\vspace*{1.0cm}
\includegraphics[width=0.48\textwidth]{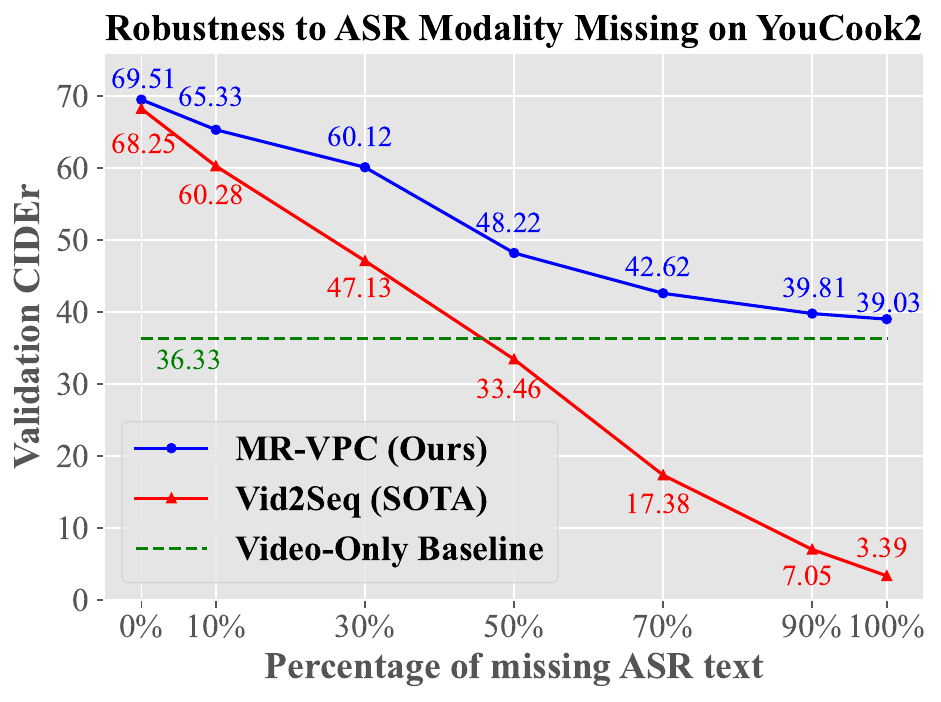}
\vspace{-0.6cm}
\caption{The performance of the previous SOTA model Vid2Seq drastically declines as the percentage of ASR text missing grows. In contrast, our \textbf{MR-VPC} consistently achieves superior performance in both modality-complete and modality-missing environments.}
\vspace{-0.5cm}
\label{fig1}
\end{figure}

\textbf{Issue-1: Under-utilization of available modalities.} 
Since a specific auxiliary modality is solely considered during training, the models fail to leverage unseen modalities that may emerge at test time.
For example, VLCap and VLTinT~\citep{yamazaki2022vlcap,yamazaki2022vltint}  cannot employ transcribed speech, which is proven extremely beneficial in Vid2Seq~\citep{yang2023vid2seq}; conversely, Vid2Seq cannot make use of event boundaries, which contain rich information about the temporal structure of videos.
\textbf{Issue-2: Vulnerability to missing modality in noisy environments.}  
The performance of these models may degrade drastically when the required auxiliary modality is absent or of low quality, which is common in real-world situations.
For instance, ~\citet{liu2021video} find that the VPC models relying on event boundaries yield significantly lower performance when the ground-truth event boundaries are replaced with learned ones.
Besides, we observe that the state-of-the-art model Vid2Seq is vulnerable to the missing of automatically transcribed speech (ASR texts)  as depicted in Figure~\ref{fig1}.~\looseness=-1

In response to \textbf{issue-1}, we design a multimodal VPC (\textbf{MVPC}) architecture to integrate the inputs from multiple modalities. 
Concretely, \textbf{MVPC} first encodes the two auxiliary modalities (\emph{i.e.}, tokenized event boundaries and transcribed speech) into a unified textual feature space using a shared text encoder. Then, the textual features are fused with the video features before entering the language decoder to generate paragraph captions. 
Further, to alleviate \textbf{issue-2}, we devise two training strategies to enhance the robustness of our model to missing modalities. 
Firstly, we simulate the absence of auxiliary modalities by randomly dropping the inputs (named \emph{DropAM}) during training. 
This approach reduces the model's reliance on auxiliary inputs and improves generalization in noisy situations. 
Second, to take full advantage of the auxiliary modalities, we propose to perform multimodal knowledge distillation~\citep{hinton2015distilling} (referred to as \emph{DistillAM}) where the model trained on modality-complete data acting as the teacher and the model operating in modality-missing situations learning as the student. 
By combining \textbf{MVPC}, \emph{DropAM} and \emph{DistillAM}, we present a \textbf{M}ultimodal noise-\textbf{R}esistant \textbf{V}ideo \textbf{P}aragraph \textbf{C}aptioning framework (\textbf{MR-VPC}).~\looseness=-1

Experimental results on two  benchmarks demonstrate the superiority of \textbf{MR-VPC} in handling both modality-complete and modality-incomplete data.
Notably, \textbf{MR-VPC} is tailored for the challenging VPC task and substantially outperforms prior robustness-oriented methods studied for classification tasks.
To our knowledge, this work pioneers formulating VPC as a multimodal learning problem with noisy inputs and presents practical solutions that enable VPC systems to utilize inputs from diverse modalities while remaining robust even when parts of them are missing.~\looseness=-1

\section{Related Work}

\paragraph{Video Paragraph Captioning (VPC)}
VPC is a widely studied video-language understanding task involving producing paragraph-level captions for long videos lasting for minutes~\citep{park2019adversarial}.
Existing VPC models commonly incorporate additional auxiliary information alongside video frames as inputs, such as transcribed speech~\citep {yang2023vid2seq} and event boundaries~\citep[etc]{zhou2018end, yamazaki2022vlcap,yamazaki2022vltint}.
\citet{liu2021video} and \citet{song2021towards} build VPC models for raw videos without event boundaries, but their models still underperform those utilizing auxiliary modalities. 
To the best of our knowledge, our work takes the first step to utilize both transcribed speech and event boundaries for VPC in an end-to-end manner, and we are the first to study the robustness of VPC models to noisy inputs with missing modalities.~\looseness=-1

\paragraph{Robustness to Missing Modality} 
As multimodal neural networks are vulnerable to missing modality~\citep{ma2022multimodal}, recent years have seen a surge of studies on enhancing model robustness on modality-incomplete data across various multimodal tasks~\citep[etc]{woo2022towards,lee2023multimodal,wei2023mmanet,yuan2023noise}. 
In terms of methodology, researchers have explored approaches such as modality fusion strategy search~\citep{ma2022multimodal}, data augmentation in the form of modality dropout~\citep{mckinzie2023robustness}, and regularization objectives~\citep{woo2022towards,mckinzie2023robustness}.
However, existing efforts are limited to relatively simple classification tasks, and model robustness to missing modality in more complex language generation tasks like VPC is yet to be explored.
We have found that simply applying the existing approaches in other tasks does not achieve satisfactory results in VPC and bridge this research gap by developing training strategies customized for VPC in our \textbf{MR-VPC} framework, which will be discussed in \S~\ref{sec:method} and \S~\ref{sec:exp}.
~\looseness=-1

\section{Methodology} \label{sec:method}

\subsection{Problem Formulation}
\begin{figure}[t]
\centering
\vspace{-0.3cm}
\includegraphics[width=0.48\textwidth]{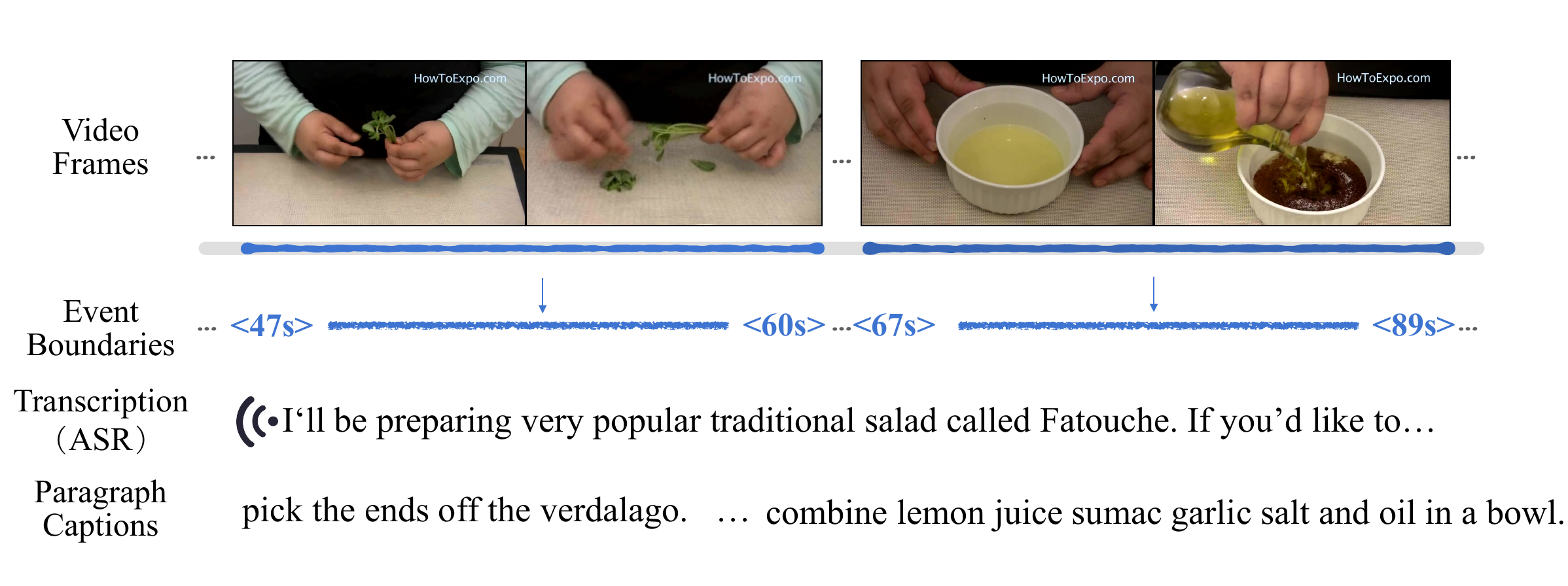}
\caption{The composition of an instance in the multimodal VPC task from the validation set of YouCook2.}
\vspace{-0.5cm}
\label{fig2}
\end{figure}

An instance  in a VPC dataset can be formulated as
$\left(V_i,A_i,E_i,C_i\right)$, where $V,A,E,C$ stand for video frames, ASR texts, event boundaries, and the caption, respectively.
An example from the YouCook2~\citep{zhou2018youcook2} dataset is illustrated in Figure~\ref{fig2}.
We assume that the video modality $V$ is always available at test time and the auxiliary modalities $A$ and $E$ are likely to be affected by noise in the wild.
Given $N_A$ and $N_E$ as the noise functions for $A$ and $E$ ($e.g.$, random missing in the context of our study on missing modality), respectively, for a model $F\left(V,A,E\right)$ trained on the clean training set $D_{\text{tr}}=\left\{ \left(V_i,A_i,E_i,C_i\right), 1 \leq i \leq n_{\text{tr}} \right\}$, where $n_{\text{tr}}$ is the size of the training data, our target is to maximize the performance on the noisy test set $D_{\text{te}}=\left\{\left(V_i,N_A\left(A_i\right), N_E\left(E_i\right),C_i\right), 1 \leq i \leq n_{\text{te}}\right\}$, where $n_{\text{te}}$ is the size of the test data. 

\begin{figure*}[thb] \centering
\includegraphics[width=0.95\textwidth]{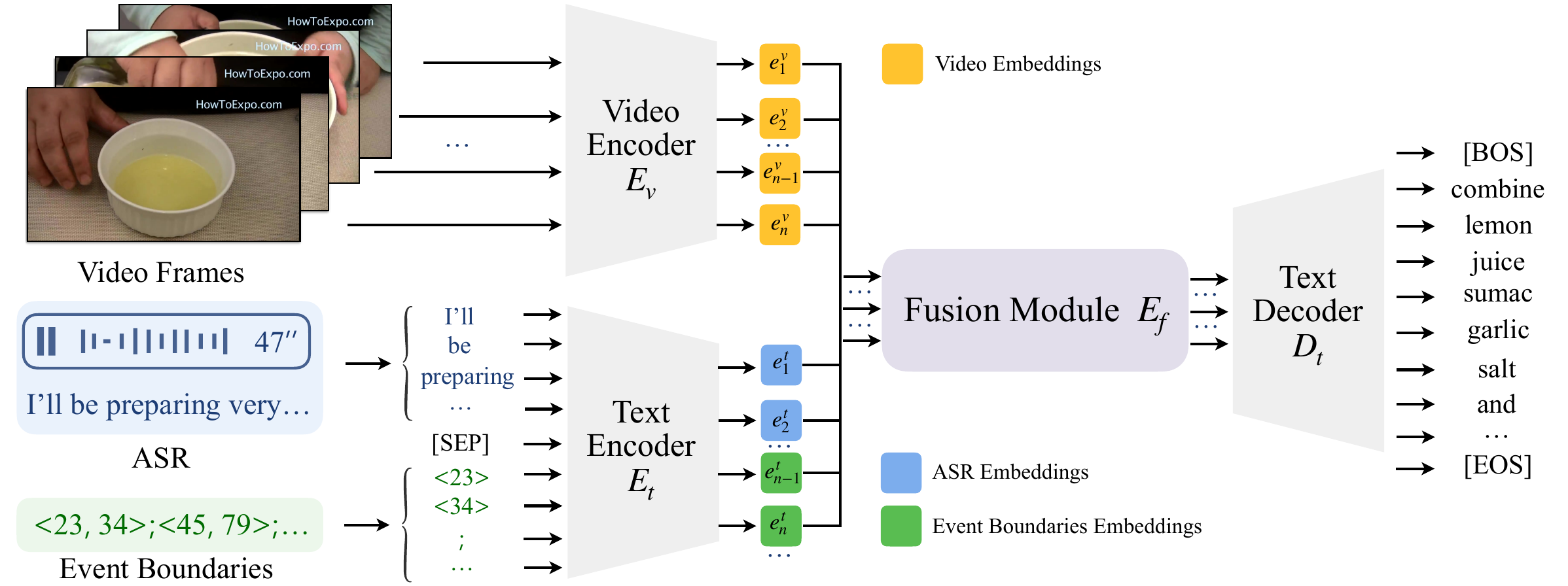}
\caption{The overview diagram of our MVPC (multimodal video paragraph captioning) framework.} 
\label{fig:model}
\vspace{-0.5cm}
\end{figure*}

\subsection{MVPC Model Framework} \label{subsec:arch}

\paragraph{Overview} 
Overall, as illustrated in Figure~\ref{fig:model}, our multimodal video paragraph captioning (MVPC) model consists of four modules: the video encoder $E_v$ to encode $V$, the text encoder $E_t$ to encode the concatenation of $A$ and $E$, a fusion module $E_f$ that merges visual and textual features, and a text decoder $D_t$ that generates the caption $C$.

\paragraph{Video Encoder}
The video encoder $E_v$ encodes the video sequence of $F$ frames 
$x_v \in \mathbb{R}^{F \times H \times W \times C} $, where $H, W$ and $C$ are the height, width, and the number of channels, respectively, and outputs the video embedding sequence $E_v\left(x_v\right) \in \mathbb{R}^{F \times d} $, where $d$ is the embedding size.
Concretely, we use a CLIP ViT-L/14~\citep{radford2021learning} image encoder to encode each frame and then feed the frame features into a 12-layer Transformer~\citep{DBLP:conf/nips/VaswaniSPUJGKP17} for temporal interaction.~\looseness=-1

\paragraph{Text Encoder}
To resolve \textbf{issue-1}, we expect  the model to be capable of modeling both $A$ and $E$ inputs end to end.
Thus before feeding $A$ and $E$ into the text encoder $E_t$, we adopt the relative time tokenization~\citep{yang2023vid2seq} to map continuous timestamps into discrete time tokens denoting the percentage progress.
Then $E_t$ transforms the concatenation of the ASR sequence and event boundary sequence $x_t$ consisting of $n$ tokens in total into the text embedding sequence $E_t\left(x_t\right) \in \mathbb{R}^{n \times d}$.~\looseness=-1

\paragraph{Fusion Module and Text Decoder} 
At the end of the workflow, the text decoder $D_t$ generates the target caption sequence in an auto-regressive manner, conditioned on the encoder embeddings produced by the fusion module $E_f$ merging $E_v\left(x_v\right)$ and  $E_t\left(x_t\right)$.
Specifically, for $E_f$,  we adopt a parameter-free concatenation operation;
for  $E_t$ and $D_t$, we employ the T5v1.1-base encoder-decoder model~\citep{raffel2020exploring}.

\paragraph{Weight Initialization}
To benefit from large-scale pretraining, we initialize the model with the Vid2Seq weight pretrained on YT-Temporal-1B~\citep{zellers2022merlot}~\footnote{Available at this \href{https://github.com/google-research/scenic/tree/main/scenic/projects/vid2seq}{link}.}.
Note that our work differs from Vid2Seq in terms of the task context and research goal.
We aim at the VPC task that generates textual paragraph-level captions $C$ from the input modalities $V,A$ and $E$, where $A$ and $E$ are likely to be missing, while Vid2Seq is originally designed for the dense video captioning task where the inputs are $V$ and $A$ (without considering missing modality) and the outputs are $C$ and $E$.
To establish a baseline for comparison, we re-implement Vid2Seq and fine-tune its pretrained weights for the VPC task (details in Appendix~\ref{app:model}). 
This allows us to evaluate the performance improvement achieved by our proposed framework.
Note that MVPC is \textbf{not} a simple extension of Vid2Seq, as our general framework to incorporate $A$ and $E$ unitedly is agnostic to the underlying structure and applies to other vision-language foundation models.~\looseness=-1

\begin{table}[t]
\centering
\resizebox{0.48\textwidth}{!}{
\begin{tabular}{@{}l|ll|ll@{}}
\toprule
\multirow{2.5}{*}{\textbf{Test Modalities}} & \multicolumn{2}{c}{\textbf{YouCook2}} & \multicolumn{2}{|c}{\textbf{ActivityNet}} \\ \cmidrule(lr){2-3} \cmidrule(lr){4-5} 
                                          & \textbf{METEOR}    & \textbf{CIDEr}   & \textbf{METEOR}     & \textbf{CIDEr}     \\ \midrule
V+E+A                                     & 23.11              & 74.13            & 14.09               & 42.29              \\ \midrule

V+A                                       & 21.05 (\textcolor{red}{-2.06})             & 59.55 (\textcolor{red}{-14.58})           & 12.24 (\textcolor{red}{-1.85})            & 29.71 (\textcolor{red}{-12.58})              \\
V+E                                       & 12.46 (\textcolor{red}{-10.65})            & 8.77 (\textcolor{red}{-65.36})          & 12.91     (\textcolor{red}{-1.18})           & 43.14   (\textcolor{green}{+0.85})            \\
V                                         & 6.79 (\textcolor{red}{-16.32})              & 3.42 (\textcolor{red}{-70.71})            & 11.64   (\textcolor{red}{-2.45})             & 26.08  (\textcolor{red}{-16.21})           \\ \bottomrule
\end{tabular}}
\vspace{-0.1cm}
\caption{The performance of the vanilla MVPC model on YouCook2 and ActicityNet Captions in different modality missing settings.}
\label{tab:youcook2_drop}
\vspace{-0.5cm}
\end{table}

\subsection{Training Strategies of MR-VPC} \label{subsec:strategy}
As the vanilla training of MVPC does not consider potential noise in the inference stage, it suffers from severe performance drops facing missing modality (\textbf{issue-2}), as shown in Table~\ref{tab:youcook2_drop}.
For instance, the absence of $A$ results in a 65.36 (88.17\% relatively) CIDEr drop on YouCook2; the missing of $E$ causes a 12.58 (29.75\% relatively) CIDEr decline on AcitivityNet.~\footnote{We find that the ASR data of ActivityNet contains little useful cues and show small negative effects, so we nullify the ASR input of ActivityNet at test time later.}
In light of this weakness, we explore the following training strategies to enhance the model's resilience to missing modality (the model trained with them is referred to as \textbf{MR-VPC} later).\looseness=-1

\subsubsection{\emph{DropAM}: Drop Auxiliary Modalities} \label{method:drop}
Since the missing modality can be viewed as a distribution shift from the training data, a fundamental idea to enhance model robustness is simulating the noise during training.
To this end, we randomly drop the auxiliary modalities $A$ and $E$ to reduce the dependence of the model on them.
Specifically, we transform the original training set $D_{\text{tr}}$ to $\hat{D}_{\text{tr}}= \left\{\left(V_i,\hat{N}_A\left(A_i\right), \hat{N}_E\left(E_i\right),C_i\right), 1 \leq i \leq n_{\text{train}}\right\}$, in which $\hat{N}_A$ and $\hat{N}_E$ are the proxy noise functions that random replace $A_i$ and $E_i$ with a default null character at probabilities $p_A$ and $p_E$, respectively:
\begin{equation} \small
\hat{N}_A\left(A_i\right) = 
\begin{cases}
    '', & p \leq p_A \\
    A_i, & p > p_A
\end{cases},
\hat{N}_E\left(E_i\right) = 
\begin{cases}
    '', & p \leq p_E \\
    E_i, & p > p_E
\end{cases},
\end{equation}
where $p$ is a random variable uniformly drawn from the range $\left[0,1\right]$.
We use $p_A=p_E=0.5$ as the value works generally well in practice. Please see the discussion about their effects in Appendix~\ref{app:drop_rate}.

\subsubsection{\emph{DistillAM}: Learning from the Teacher with Modality-Complete Data}
Solely applying \emph{DropAM} turns the model training into a multitask learning process involving subtasks with different input conditions, which possibly adds to the learning difficulty and compromises the performance on modality-complete data.
Therefore, we resort to knowledge distillation~\citep{hinton2015distilling}, a learning paradigm that transfers the knowledge from teacher models with better conditions, such as more training data and a larger number of parameters, to student models without these advantages.
In our problem,  we consider the vanilla MVPC model trained on the modality-complete training set $D_{\text{tr}}$ as the teacher model $F_t$, and our goal is to transfer the knowledge learned by $F_t$ to the \textbf{MR-VPC} model that likely faces missing modality as the student model $F_s$.
In early trials, we have found that distilling from word-level logits (WordKD) achieves limited performance gains in our task.
Therefore, inspired by the sequence-level knowledge distillation (SeqKD)~\citep{kim2016sequence} studied in machine translation, we create a new training set $D_{\text{kd}}$ by replacing the ground-truth caption $C$ with the predictions given by $F_t$ based on the modality-complete data: 
\begin{equation} \small
    D_{\text{kd}} = \left\{\left(V_i,A_i,E_i, F_t\left(V_i,A_i,E_i\right)\right), 1 \leq i \leq n_{\text{tr}}\right\},
\end{equation}
and then construct the augmented training set $D_{\text{aug}}= D_{\text{tr}} \bigcup D_{\text{kd}}$ by merging $D_{\text{kd}}$ and the original training data $D_{\text{tr}}$. 
It is notable that this procedure named \emph{DistillAM}  is orthogonal to the noise simulation process  
\emph{DropAM} in \S~\ref{method:drop}, so they can be applied together, \emph{i.e.}, the random noise can be injected into the augmented training data $D_{\text{aug}}$ in the training phase in the way stated in \S~\ref{method:drop}.


\subsubsection{Connection to Prior Strategies for Multimodal Classification Tasks}
Although MASD~\citep{mckinzie2023robustness}, the state-of-the-art approach to enhance model robustness to missing modality in classification problems, also takes the form of modality dropout and knowledge distillation, it differs from our solutions in essence.
Concretely, MASD performs self-distillation, namely aligning the predicted probabilities on modality-complete and modalities-incomplete data output by the same model under training.
In contrast, we use a fixed teacher model trained on modality-complete data, which facilitates the efficient learning of the student model in the challenging VPC task.
We will show the advantage of our MR-VPC over MASD and its variant MASD+Wise-FT~\citep{mckinzie2023robustness} in \S~\ref{subsubsec:comp_sota}.~\looseness=-1

\section{Experiments}  \label{sec:exp}

\subsection{Experimental Setup} \label{sunsec:exp_setup}

\paragraph{Evaluation Protocol}
Following ~\citet{yang2023vid2seq}, we use CIDEr (C)~\citep{vedantam2015cider} and METEOR (M)~\citep{banerjee2005meteor} metrics to evaluate the accuracy of generated captions.
For measuring diversity, we use 4-gram repetition (R@4)~\citep{xiong2018move} following \citet{liu2021video} and \citet{yamazaki2022vlcap,yamazaki2022vltint}.
Besides these metrics based on n-gram matching commonly used in previous works, we also report advanced model-based metrics in \S~\ref{subsec:model_metrics}.\looseness=-1

\paragraph{Benchmarks} We conduct main experiments on YouCook2~\citep{zhou2018youcook2} and ActivityNet Captions~\citep{krishna2017dense}, two widely studied VPC benchmarks containing paragraph-level captions and annotated event boundaries.
We report the evaluation metrics on the validation set of YouCook2 and the \textit{as-test} split of ActivityNet Captions (see Appendix~\ref{app:datasets} for details).

\paragraph{Acquisition of ASR Data}
For ActivityNet Captions, we adopt the ASR data provided by ~\citet{MDVC_Iashin_2020} from the YouTube ASR system.
For YouCook2, we obtain the ASR data using the \textit{whisper-timestamped} tool~\citep{lintoai2023whispertimestamped} based on Whisper~\citep{whisper} (the \textit{small.en} model with 244M parameters) and dynamic time warping~\citep{JSSv031i07}.

\paragraph{Model Training and Inference}
We train the model for 40 epochs on YouCook2 and 20 epochs on ActivityNet Captions using a batch size of 32.
The model is trained with the Adam~\citep{adam} optimizer to minimize cross-entropy loss with an initial learning rate of 2e-4 with cosine annealing.
For training efficiency, we freeze the image encoder in our experiments unless otherwise mentioned, so the number of trainable parameters is 314M.
The weight decay is 5e-2 and we clip the maximum norm of the gradient to 1.0. 
We uniformly sample 100 frames at resolution 224$\times$224 pixels for the video input and the ASR text sequence is truncated at the max length of 1000.
Temporally consistent random spatial augmentation~\citep{qian2021spatiotemporal} is applied.
The inference beam search size is 4 and the repetition penalty is 1.2.
See more details in Appendix~\ref{app:model}.\looseness=-1
 
\paragraph{Evaluation Settings}

We mainly report results in three representative test settings:
(1) \textbf{the modality-complete setting} where the auxiliary modalities $A$ and $E$ are not affected by any noise; 
(2) \textbf{the video-only setting} where both $A$ and $E$ are missing, which is a harsh but realistic setting (in the real world, most users do not enter the video's event boundaries $E$; $A$ is also possibly missing, \textit{e.g.}, when the ASR system does not support the conversation language);
(3) \textbf{the random-missing setting} where $A$ and $E$ are both randomly missing at the probability of 50\% independently.

\paragraph{Baselines} 
We compare our models with a wide array of baselines and categorize them according to the input modalities in their original settings:

$\bullet$ \textbf{V:} The Vid2Seq model finetuned on only the video modality, named Vid2Seq (V); SoftNMS~\citep{bodla2017soft}, ESGN~\citep{mun2019streamlined}, Memory Transformer~\citep{song2021towards}, and VPCSum~\citep{liu2021video}; MART, MART$^{\text{COOT}}$,  Vanilla Transformer, and Transformer-XL. 
The last four models use event boundaries generated by ESGN at test time as done in ~\citet{liu2021video}.~\looseness=-1

$\bullet$ \textbf{V+E:} VLTinT~\citep{yamazaki2022vltint}, VLCap~\citep{yamazaki2022vlcap}, MART~\citep{lei2020mart}, MART$^{\text{COOT}}$~\citep{ging2020coot}, Vanilla Transformer~\citep{zhou2018end}, and Transformer-XL~\citep{dai2019transformer}.

$\bullet$ \textbf{V+A:} Vid2Seq~\citep{yang2023vid2seq}.~\looseness=-1

\subsection{Results and Analysis} \label{subsec:main_results}

\subsubsection{Comparing MVPC and MR-VPC}

\begin{table}[t]
\centering
\resizebox{0.48\textwidth}{!}{
\begin{tabular}{@{}l|cc|cccc@{}}
\toprule
\multirow{2.5}{*}{\textbf{Model}} & \multicolumn{2}{c}{\textbf{Training Strategies}} &  \multicolumn{4}{|c}{\textbf{Test Modalities}}      \\
\cmidrule(lr){2-3} \cmidrule(lr){4-7}
 & \emph{\textbf{DropAM}} & \emph{\textbf{DistillAM}} & \textbf{V+E+A} & \textbf{V+E} & \textbf{V+A} & \textbf{V} \\ \midrule
MVPC           &    \XSolidBrush            &      \XSolidBrush            &            \textbf{74.13}/\textbf{23.11}    &   8.77/12.46           &     59.55/21.05         &    3.42/6.79        \\
-              &       \Checkmark           &   \XSolidBrush              &  60.40/22.67              &    35.17/16.94          &  64.87/22.54            &       36.73/16.53     \\
MR-VPC         &    \Checkmark  &     \Checkmark                &      69.51/22.83          &     \textbf{39.03}/\textbf{16.97}        &     \textbf{69.37}/\textbf{22.59}        &   \textbf{38.37}/\textbf{16.86}        \\ \bottomrule
\end{tabular}}
\caption{The effect of our training strategies with different available modalities at test time on the YouCook2 dataset. CIDEr / METEOR metrics are reported.}
\label{tab:ablation}
\vspace{-0.5cm}
\end{table}

\paragraph{Our training strategies remarkably boost the model's robustness to missing modality while maintaining the performance in the modality-complete setting.}
Before comparing our model with baselines, we first examine the effectiveness of our training strategies described in \S~\ref{subsec:strategy}.
As the results displayed in Table~\ref{tab:ablation}, the vanilla MVPC model without these training strategies is extremely susceptible to missing modality at test time, but the MR-VPC model equipped with these techniques shows substantially improved robustness to missing modality with only minimal performance sacrifice on the modality-complete test data.
For instance, MVPC disastrously fails in the video-only setting (the CIDEr falls to 3.42), while MR-VPC yields a CIDEr value of 38.37.
We also affirm the validity of each strategy by comparing MR-VPC with the model trained with only the \emph{DropAM} strategy (the last two rows of Table~\ref{tab:ablation}).
As shown, although \emph{DropAM} boosts the model robustness on modality-incomplete data, it significantly hurts the performance on modality-complete data (the CIDEr declines from 74.13 to 60.40); \emph{DistillAM} not only further advances the robustness to missing modality, but also help preserve the performance in the modality-complete setting, as it raises the CIDEr metric to 69.51.~\looseness=-1

\begin{table}[t] \small
\centering
\begin{tabular}{@{}lcc@{}}
\toprule
\textbf{BERTScore$\uparrow$} & \textbf{YouCook2} & \textbf{ActivityNet} \\ \midrule
MVPC               &   82.37                &      91.72                \\
MR-VPC              &    \textbf{91.30}               &   \textbf{94.16}                 \\  \bottomrule
\end{tabular}
\caption{The average BERTScore similarities between captions generated in modality-complete and video-only test scenarios.}
\label{table:bert_score_comparison}
\vspace{-0.5cm}
\end{table}

\begin{figure}[t]
    \centering
    \begin{subfigure}[Captions generated by the vanilla MVPC model.]{ \label{subfig:tsne_vanilla}
        \includegraphics[width=0.45\textwidth]{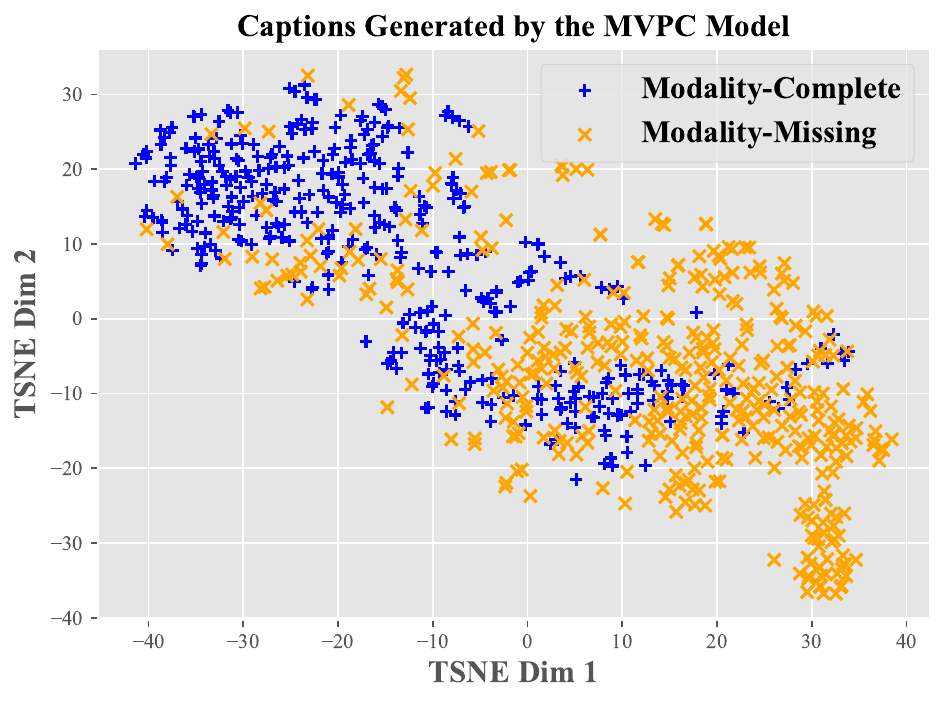}
    }
    \end{subfigure} 
    
    \begin{subfigure}[Captions generated by the MR-VPC model.]{ 
    \label{subfig:tsne_robust}
        \includegraphics[width=0.45\textwidth]{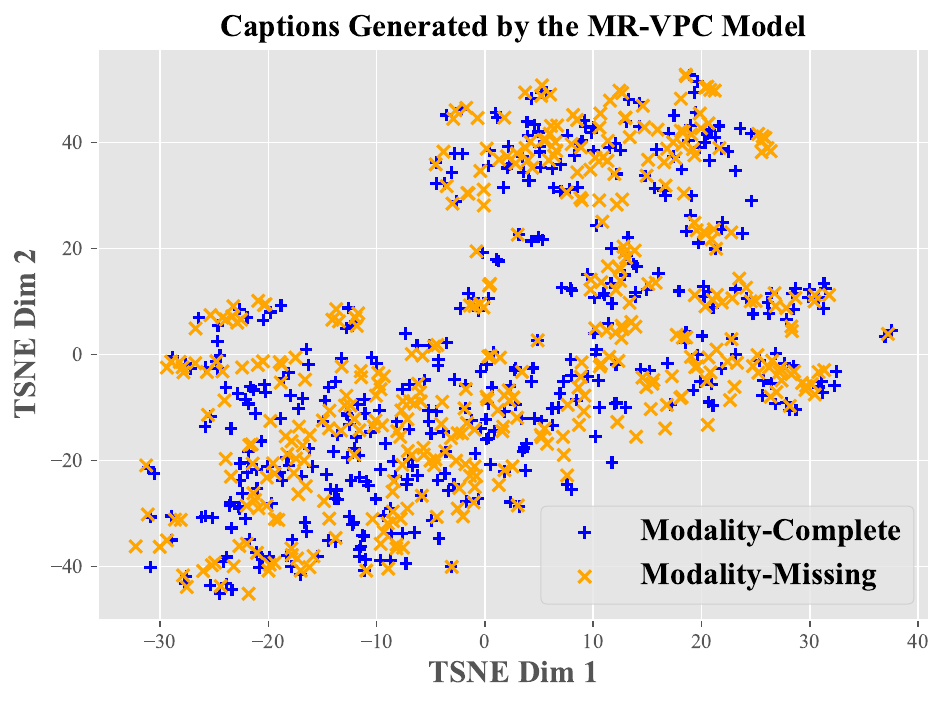}
    }
    \end{subfigure}
    \vspace{-0.2cm}
    \caption{Visualization of the SimCSE embeddings of the captions generated under \textcolor{blue}{modality-complete} and \textcolor{orange}{modality-missing} (video-only) scenarios.}
    \label{fig:tsne}
    \vspace{-0.5cm}
\end{figure}

\paragraph{MR-VPC shows higher prediction consistency between modality-complete and modality-missing scenarios.} To intuitively understand the impact of our training strategies, we compare the BERTScore~\citep{zhangbertscore} similarities between the captions generated on modality-complete and video-only data by the vanilla MVPC and MR-VPC models.
As listed in Table~\ref{table:bert_score_comparison}, MR-VPC exhibits substantially higher similarity scores, which indicates that it is capable of generating more consistent predictions, regardless of the availability of auxiliary modalities.
Furthermore, we visualize the SimCSE embeddings~\citep{gao2021simcse}~\footnote{We use the unsup-simcse-roberta-large model.} of the generated captions on YouCook2 using t-SNE~\citep{van2008visualizing} in Figure~\ref{fig:tsne}, where we observe that the captions generated by MVPC form two distinct clusters depending on whether modality-missing occurs, but those produced by MR-VPC appear in pairs and seem hard to distinguish based on the test scenario. 
The visualization further proves that \emph{DropAM} and \emph{DistillAM} contribute to the consistency of the predictions.\looseness=-1

\subsubsection{Comparison with Advanced Systems} \label{subsubsec:comp_sota}

\begin{table}[t]
\centering
\resizebox{0.48\textwidth}{!}{
\begin{tabular}{@{}l|ccc|ccc@{}}
\toprule
\multirow{2.5}{*}{\textbf{Model}} & \multicolumn{3}{c}{\textbf{YouCook2}}                                                            & \multicolumn{3}{|c}{\textbf{ActivityNet}}                                                      \\  \cmidrule(lr){2-4} \cmidrule(lr){5-7} & \textbf{C $\uparrow$} & \textbf{M $\uparrow$} & \textbf{R@4 $\downarrow$ }  & \textbf{C $\uparrow$} & \textbf{M $\uparrow$} & \textbf{R@4 $\downarrow$ } \\ \midrule
      MVPC (\emph{Ours})    &     \textbf{74.13}      &    \textbf{23.11}           &         0.82                       &                    \textbf{43.14}           &        13.91                        &     0.67                          \\
      MR-VPC  (\emph{Ours})                         &    69.51                            &                 22.83               &               \textbf{0.57}                 &        41.01                        &      13.84                         &         \textbf{0.51}              \\    \midrule
    \emph{Baselines} & & & & & & \\
 Vid2Seq       &        68.25                     &  23.01                             &             0.75                   &             30.77                   &                       12.51         &    0.82                            \\
 Vid2Seq (V) &  36.33 & 16.79 & 0.79 & 28.87 & 12.38 & 0.57 \\
                 VLTinT               &      48.70                          &                     17.94           &        4.29                        &                  31.13              &       \textbf{17.97}                         &              4.75                  \\
     VLCap                           &  49.41                              &                    17.95            &      5.16                          &                        30.29        &        17.48                        &    4.18                           \\
                 MART               &        35.74                       &    15.90                            &        4.39                   &                  22.16               &      15.57                          &                      5.44         \\
                 MART$^{\text{COOT}}$               &  46.06                              &              18.17                  &     6.30                           &    28.19                            &                      15.99          &     6.64              \\
Vanilla Trans.  & 38.00 & 11.55 & - & 21.33 & 15.54 & 7.45 \\
Memory Trans. &  - & - & - & 26.55 & 15.64 & 2.75 \\
Trans.-XL & 26.40 & 14.80 & - & 21.71 & 14.91 & 8.79 \\   
VPCSum & 23.92 & 15.11 & 0.65 & 24.33 & 15.84 & 1.54 \\
\bottomrule             
\end{tabular}}
\caption{Evaluation results under the modality-complete setting. $\uparrow$ indicates larger is better and $\downarrow$ indicates lower is better. The best result is highlighted in \textbf{bold}.}
\label{tab:main_results_complete}
\vspace{-0.3cm}
\end{table}

\paragraph{Our MVPC and MR-VPC obtain superior performance in the modality-complete setting.}
We present the evaluation results in the modality-complete setting in Table~\ref{tab:main_results_complete} and observe that our models markedly advance the state-of-the-art on most metrics.
In terms of captioning accuracy, we elevate the CIDEr metric from 68.25 (Vid2Seq) to 74.13 on YouCook2 and from 31.13 (VLTinT) to 43.14 on ActivityNet; regarding diversity, we achieve the lowest R@4 repetition scores below 1.0.
These results support the necessity to fully leverage the auxiliary modalities $A$ and $E$ (\textbf{issue-1}) and the effectiveness of our MVPC model frameowork.
We notice that VLTinT and some earlier baselines do better in terms of METEOR on AcitivyNet than Vid2Seq and our models, but we contend that ours and Vid2Seq are better models for two reasons: (1) CIDEr is a more reasonable metric because it accounts for the importance of different n-grams and has shown higher consistency with human evaluation~\citep{shi2022emscore}; (2) model-based metrics in \S~\ref{subsec:model_metrics} and human study results in \S~\ref{subsec:human} further corroborate the advantages of our models.~\looseness=-1

\begin{table}[t]
\centering
\resizebox{0.48\textwidth}{!}{
\begin{tabular}{@{}l|ccc|ccc@{}}
\toprule
\multirow{2.5}{*}{\textbf{Model}} & \multicolumn{3}{c}{\textbf{YouCook2}}                                                            & \multicolumn{3}{|c}{\textbf{ActivityNet}}                                                      \\  \cmidrule(lr){2-4} \cmidrule(lr){5-7} &  \textbf{C $\uparrow$} & \textbf{M $\uparrow$} & \textbf{R@4 $\downarrow$ }  & \textbf{C $\uparrow$} & \textbf{M $\uparrow$} & \textbf{R@4 $\downarrow$ } \\ \midrule
MVPC (\emph{Ours})   & 3.42  & 6.79  & 2.31  & 26.08   & 11.64  & 0.60 \\
MR-VPC  (\emph{Ours})   &  \textbf{38.37} & \textbf{16.86} & \textbf{0.57}  & \textbf{31.37} & 12.06 & \textbf{0.58}  \\    \midrule
\emph{Baselines} & & & & & & \\
Vid2Seq  & 3.39  & 6.81  & 2.80 & 30.01  & 12.18 & 0.73 \\
Vid2Seq (V) &  36.33 & 16.79 & 0.79 & 28.87 & 12.38 & \textbf{0.58} \\
Memory Trans. &  - & - & - & 26.55 & 15.64 & 2.75 \\ 
VPCSum & 23.92 & 15.11 & 0.65 & 24.33 & \textbf{15.84} & 1.54 \\
SoftNMS & 18.18 & 13.67 & 4.94 & 22.58 & 14.93 & 10.17 \\ 
ESGN & 21.85 & 15.74 & 6.51 & 17.01 & 13.37 & 4.94 \\
Vanilla Trans. & 20.95 & 15.11 & 7.04 & 16.88 & 13.37 & 2.85 \\
Trans.XL & 14.24 & 12.67 & 3.20 & 20.73 & 14.89 & 7.45 \\
MART & 16.56 & 13.44 & 4.63 & 20.16 & 14.94 & 6.09 \\
COOT & 19.67 & 14.21 & 5.99 & 21.83 & 14.67 & 1.54 \\
\bottomrule             
\end{tabular}}
\vspace{-0.1cm}
\caption{Evaluation results under the video-only setting.}
\label{tab:main_results_video_only}
\vspace{-0.2cm}
\end{table}

\begin{table}[t]
\centering
\resizebox{0.48\textwidth}{!}{
\begin{tabular}{@{}l|ccc|ccc@{}}
\toprule
\multirow{2.5}{*}{\textbf{Model}} & \multicolumn{3}{c}{\textbf{YouCook2}}                                                            & \multicolumn{3}{|c}{\textbf{ActivityNet}}                                                      \\  \cmidrule(lr){2-4} \cmidrule(lr){5-7} &  \textbf{C $\uparrow$} & \textbf{M $\uparrow$} & \textbf{R@4 $\downarrow$ }  & \textbf{C $\uparrow$} & \textbf{M $\uparrow$} & \textbf{R@4 $\downarrow$ } \\ \midrule
MVPC (\emph{Ours})   & 33.31  & 15.70 & 1.55  & 33.55    &  12.86  & 0.59 \\ 
MR-VPC  (\emph{Ours})   &  \textbf{51.13} & \textbf{20.15}  &  0.74 & \textbf{37.05} & 13.01 &  \textbf{0.56}  \\    \midrule
\emph{Baselines} &  & & & & & \\
Vid2Seq  & 33.46  & 14.19 & 1.46  & 29.93  & 12.48  & 0.75  \\
Vid2Seq (V) &  36.33 & 16.79 & 0.79 & 28.87 & 12.38 & 0.58 \\
VPCSum & 23.92 & 15.11 & \textbf{0.65} & 24.33 & \textbf{15.84} & 1.54 \\
\bottomrule             
\end{tabular}}
\vspace{-0.2cm}
\caption{Results under the random-missing setting.~\looseness=-1}
\label{tab:main_results_random}
\vspace{-0.5cm}
\end{table}

\paragraph{Our MR-VPC model performs significantly better in modality-missing settings than previous SOTA models.}
Given the figures displayed in Table~\ref{tab:main_results_video_only} and Table~\ref{tab:main_results_random}, MR-VPC yields the best performance in the video-only and random-missing setting with substantial margins over baselines including those specially trained for the video-only setting such as Vid2Seq (V)~\citep{yang2023vid2seq}, VPCSum~\citep{liu2021video}, and Memory Transformer~\citep{song2021towards}. 
This suggests that MR-VPC fulfills our objective of developing a robust VPC model capable of leveraging available auxiliary modalities while maintaining robustness even when they are missing in real-world scenarios.

\begin{table}[t] \small
\centering
\resizebox{0.48\textwidth}{!}{
\begin{tabular}{@{}l|ccc@{}} 
\toprule
\textbf{Model} & \textbf{CIDEr} & \textbf{BERTScore} & \textbf{BARTScore} \\ \midrule
MVPC           &     6.79           &  87.08    & -4.56              \\
MR-VPC         &     \textbf{8.74}           &   \textbf{87.22}            &  \textbf{-4.47}          \\
Vid2Seq        &    4.74            &   86.83            &    -4.62           \\
Vid2Seq (V)    &     6.01           &   87.00            &      -4.48         \\ \bottomrule
\end{tabular}}
\caption{Zero-shot evaluation results on Charades (the model weights are trained on ActivityNet Captions).}
\label{tab:charades}
\vspace{-0.3cm}
\end{table}
\paragraph{Our MR-VPC  shows the best cross-dataset generalization performance on the video-only Charades dataset.} 
To further examine the cross-dataset generalization capability, we assess the models trained on ActivityNet Captions on the test set of the Charades~\citep{sigurdsson2016charades}, where only the video modality is available.
As the results listed in Table~\ref{tab:charades}, MR-VPC outperforms baselines in the zero-shot scenario where domain shift and missing modality occur simultaneously, further validating the strength of our approach.\looseness=-1

\paragraph{Our MR-VPC beats the SOTA robustness-oriented training methods in classification problems.}
As shown in Table~\ref{tab:masd_comparsion}, MR-VPC remarkably outperforms the state-of-the-art solutions towards robustness to missing modality in classification problems, \textit{i.e.}, MASD and MASD+Wise-FT~\citep{mckinzie2023robustness}.
This illustrates that our customized approaches for the VPC task make significant strides compared to simply incorporating existing techniques studied for other tasks previously.
Besides, we observe that replacing the SeqKD with Word-KD leads to significant performance drops in all scenarios, which supports the rationality of using SeqKD in our \emph{DistillAM} component.

\begin{table}[t] 
\centering
\resizebox{0.48\textwidth}{!}{
\begin{tabular}{@{}l|cccc|c@{}}
\toprule
\multirow{2.5}{*}{\textbf{Method}} &   \multicolumn{4}{c|}{\textbf{Test Modalities}} &  \multirow{2.5}{*}{\textbf{Avg.}}    \\
 \cmidrule(lr){2-5}
 & \textbf{V+E+A} & \textbf{V+E} & \textbf{V+A} & \textbf{V} \\ \midrule
 WordKD  &      64.50           &    30.62       &  65.33   &   27.21 & 46.92   \\
MASD           &      67.95  &  32.98   &     68.72     &   33.47  &   50.78  \\
MASD+WiSE-FT             &     68.90            &  34.96         &  \textbf{69.54} &   32.54  & 51.49 \\
MR-VPC (Ours)        &     \textbf{69.51}      &    \textbf{39.03}    &   69.37    &  \textbf{38.37}   &  \textbf{54.07} \\
\bottomrule
\end{tabular}}
\caption{Comparison with other robustness-oriented methods with different available modalities at test time on YouCook2. CIDEr metrics are reported.}
\label{tab:masd_comparsion}
\vspace{-0.5cm}
\end{table}

\begin{table*}[t] \small
\centering
\resizebox{0.95\textwidth}{!}{
\begin{tabular}{@{}l|ccc|ccccc@{}} 
\toprule
\multirow{2.5}{*}{\textbf{Model}} & \multicolumn{3}{c}{\textbf{YouCook2}}                                      & \multicolumn{5}{|c}{\textbf{ActivityNet Captions}}                                   \\  \cmidrule(lr){2-4} \cmidrule(l){5-9} 
                                & \textbf{PPL $\downarrow$} & \textbf{BERT $\uparrow$} & \textbf{BART $\uparrow$} & \textbf{PPL $\downarrow$} & \textbf{BERT $\uparrow$} & \textbf{BART $\uparrow$} & \textbf{EMS $\uparrow$} & \textbf{EMS$_{\textbf{ref}}$ $\uparrow$} \\ \midrule
VLTinT~\citep{yamazaki2022vltint}                        &    21.99          &   89.01            &   -3.91            &     30.97      &  88.03             &   -3.94           &     28.94          &         36.88                     \\ 
Vid2Seq~\citep{yang2023vid2seq}                     &     15.89         &    \textbf{90.58}         &    \textbf{-3.08}          &    24.68       & 88.71             & -3.78        &  \textbf{29.54}           & 36.99                   \\  
MVPC (Ours)                           &  15.50            &    90.56           &        \textbf{-3.08}       &     18.77         &     \textbf{88.98}         &    \textbf{-3.56}          &           29.37    &    \textbf{37.21}                    \\
MR-VPC  (Ours)                        &       \textbf{15.11}       &    89.51           &      -3.49         &     \textbf{17.17}         &   88.85           &   -3.58    &      29.10      &          36.90                \\
\bottomrule
\end{tabular}}
\vspace{-0.10cm}
\caption{The model-based metrics evaluated under the modality-complete setting. $\uparrow$ indicates higher is  better and $\downarrow$ indicates lower is better. We highlight the best model in \textbf{bold}. We do not report EMScore on YouCook2 as the captions of YouCook2 are longer than the max length limit of CLIP, the backbone of the EMScore metric.\looseness=-1}
\label{tab:model_based}
\end{table*}
\begin{table*}[h!]
\centering
\resizebox{0.95\textwidth}{!}{
\begin{tabular}{@{}l|ccc|ccc|ccc|ccc|ccc@{}} \toprule
 \textbf{Noise Type} & \multicolumn{3}{c}{\textbf{Low-Quality ASR}} & \multicolumn{3}{|c}{\textbf{ASR Sentence Deletion}} & \multicolumn{3}{|c}{\textbf{Event Deletion}} & \multicolumn{3}{|c}{\textbf{Boundary Perturbation}} & \multicolumn{3}{|c}{\textbf{Generated Boundary}} \\ \midrule
 \textbf{Metric} &     \textbf{CIDEr}     &   \textbf{BERT}   &   \textbf{BART}    &      \textbf{CIDEr}  &  \textbf{BERT}    &   \textbf{BART}      &  \textbf{CIDEr}     &   \textbf{BERT}   &    \textbf{BART}    &   \textbf{CIDEr}   &   \textbf{BERT}   &   \textbf{BART} &   \textbf{CIDEr}   &   \textbf{BERT}   &   \textbf{BART}  \\ \midrule
  Vid2Seq  &  60.39     &  90.35      &  -3.12     & 48.01         &  89.62     &   -3.31     &  68.25      &  90.58    & -3.08       &      68.25 & 90.58  &  -3.08  &     \textbf{68.25} & 90.58  &  -3.08  \\
  MVPC (Ours)  & 59.58    &  90.36      & -3.13      &  48.95     & 89.66      &    -3.29   &  63.43    &  90.54      & -3.11   & \textbf{72.60} & 90.57 & -3.07  & 61.71	 & 90.58 &	-3.07    \\
 MR-VPC (Ours) &  \textbf{63.69}     & \textbf{90.63}       &  \textbf{-3.08}   & \textbf{53.59} &   \textbf{90.04}    & \textbf{-3.24} & \textbf{70.72}  & \textbf{90.85}  & \textbf{-3.02}  &  69.11    & \textbf{90.86}    & \textbf{-3.03}  & 67.02 &	\textbf{90.84}	& \textbf{-3.03}   \\
 \bottomrule
\end{tabular}}
\vspace{-0.2cm}
\caption{The evaluation results under five forms of noise in auxiliary modalities.}
\vspace{-0.4cm}
\label{tab:other_noise}
\end{table*}
\begin{table}[h] \small
\centering 
\begin{tabular}{@{}cccc@{}} 
\toprule
\multirow{2.5}{*}{Group1} & MVPC   & VLTinT & Equal \\ \cmidrule(l){2-4}
                        &   56.0\%     &  20.7\%       &    23.3\%   \\ \midrule
\multirow{2.5}{*}{Group2} & MR-VPC & VLTinT & Equal \\ \cmidrule(l){2-4}
                        &    56.0\%     &   18.7\%      &   25.3\%  \\  \bottomrule
\end{tabular}
\vspace{-0.2cm}
\caption{The average percentage of human preferences.}
\label{tab:human}
\vspace{-0.3cm}
\end{table}

\subsection{Qualitative Results}
Besides the above quantitative results, we provide qualitative evidence to support the superiority of our models.
First, we find that MVPC and Vid2Seq tend to produce degenerated captions in the modality-missing setting, whereas the prediction of MR-VPC remains almost unchanged, as exemplified by the instance given in Table~\ref{tab:pred_case} in Appendix~\ref{app:qual}.
Moreover, even in the modality-complete setting, the Vid2Seq and VLTinT baselines often predict concepts that are not present in the video; in contrast, our MVPC and MR-VPC model produces fewer such hallucinations, as illustrated in Figure~\ref{fig:demo1} in Appendix~\ref{app:qual}.~\looseness=-1

\section{Further Evaluation}

\subsection{Evaluation with Model-Based Metrics} \label{subsec:model_metrics}

Besides the n-gram-based metrics reported in \S~\ref{subsec:main_results}, we further compare our models with competitive baselines (Vid2Seq and VLTinT) using the following model-based metrics (details in Appendix~\ref{app:metrics}), as they align better with human preference~\citep{shi2022emscore}: (1) \textbf{Perplexity (PPL)} for fluency; (2) \textbf{BERTScore }~\citep{zhangbertscore} and \textbf{BARTScore}~\citep{yuan2021bartscore} measuring prediction-reference similarity; (3) \textbf{EMScore}~\citep{shi2022emscore} for the matching extent of the prediction and the video frames and its extension \textbf{EMS$_{\textbf{ref}}$}. 
We present the results in Table~\ref{tab:model_based} and find that our MVPC and MR-VPC obtain the best performance across most of these metrics. 
Notably, although VLTinT reaches the highest METEOR on ActivityNet, it falls behind our models and Vid2Seq on these metrics.
We will further show the advantage of our models through human evaluation in \S~\ref{subsec:human}.\looseness=-1

\subsection{Generalization on Other Forms of Noise} \label{subsec:other_noise}
Besides completely missing, the auxiliary modalities in the real world may also be affected by other weaker forms of noise, such as variations in ASR quality between the training and test phases.
We further test our models and VidSeq under five types of noise: lower ASR quality and sentence deletion for $A$; event deletion, boundary perturbation, and generated boundaries for $E$ (details in Appendix~\ref{app:noise}). 
We present the results in Table~\ref{tab:other_noise} and see that although these forms of noise are not seen during training, our MR-VPC shows the best robustness in most cases, which again substantiates the generalizability of our training strategies.
We believe that we will achieve even better robustness to these types of noises if we consider them in the choice of the proxy noise functions $\hat{N}_A$ and $\hat{N}_E$ in \emph{DropAM}.

\subsection{Human Evaluation} \label{subsec:human}
We conduct two groups of human evaluation, in which three annotators compare the captions
generated by VLTinT and MVPC (or MR-VPC) in the modality-complete setting for 50 randomly sampled videos from the AcitivityNet Captions test set.
They need to choose a caption showing higher consistency with the video content or mark that two captions are equally good (details in Appendix~\ref{app:human}).
As shown in Table~\ref{tab:human}, our MVPC and MR-VPC significantly surpass VLTinT in pair-wise comparison, which again proves their superiority.\looseness=-1
\vspace{-0.1cm}
\section{Conclusion}
We present MR-VPC, a multimodal video paragraph captioning model capable of utilizing three input modalities (video, transcribed speech, and event boundaries) and keeping robust in the presence of missing modality.
The MR-VPC framework comprises two key contributions: (1) the MVPC architecture, which seamlessly processes inputs from all three modalities in an end-to-end manner; (2) the incorporation of two training techniques, \emph{DropAM} and \emph{DistillAM}, which enhance the model's robustness when faced with missing modality.
Through exhaustive experimental evaluation on YouCook2 and ActivityNet Captions datasets, we demonstrate the superiority of MR-VPC in various test scenarios, highlighting its practicality and efficacy in addressing the challenges of video paragraph captioning in real-world settings.~\looseness=-1

\section*{Limitations}
We discuss the limitations of our work as follows.
(1) Despite the outstanding performance of MR-VPC in modality-missing settings, it slightly lags behind our vanilla MVPC in the modality-complete setting.
This is comprehensible because the optimization of the regularization targets introduced in \emph{DropAM} and \emph{DistillAM} may conflict with the learning on modality-complete data to some extent.
We will conduct more explorations to reduce this gap.
(2) We primarily study the absence (discussed in most of the main text) and other forms of noise (studied in \S~\ref{subsec:other_noise}) in two main auxiliary modalities, namely transcribed speech and event boundaries, which do not cover all possible harsh test conditions in the wild.
For future work, we intend to investigate the robustness of VPC models to other forms of data noise, such as video frame blurring, for a more comprehensive evaluation.\looseness=-1

\section*{Ethics Statement}

We believe that our proposal would contribute to the robustness and security of video captioning systems deployed in the open-world environment, as the absence and quality reduction of auxiliary modalities are common in practice.
Our proposal also applies to other multimodal natural language generation tasks, \emph{e.g.}, multimodal machine translation, on which we plan to conduct more studies in the future.
Moreover, all pretrained models used in this work are publicly available, ensuring transparency and accessibility.
Although we do not expect any direct negative consequences resulting from this paper, we hope to continue to build on our MR-VPC framework and develop stronger and safer multimodal VPC models in our future work.


\bibliography{custom}

\appendix

\section{Dataset Statistics} \label{app:datasets}
We conduct main experiments on YouCook2~\cite {zhou2018youcook2} and ActivityNet Captions~\citep{krishna2017dense}.
YouCook2 consists of 1,333 videos in the training set and 457 ones in the validation set.
Each instance in YouCook2 has 7.7 event segments on average.
ActivityNet Captions comprises 10,009 samples in the training set and 4,917 ones in the original validation set.
Following the practice of \citet{lei2020mart} and most of the baselines, we split the validation set to the \textit{as-val} set of 2,460 videos and the \textit{as-test} split of 2,457 videos.
Each sample in ActicityNet Captions has 3.65 event segments on average.
The average video length is 2.0 minutes in ActivityNet Captions and 5.3 minutes in YouCook2.
Also, we test the cross-dataset performance on the test set of the Charades~\citep{sigurdsson2016charades} dataset consisting of 1,760 videos.
The average video length of Charades is 30s.~\looseness=-1

\section{More Implementation Details} \label{app:model}

\paragraph{Vid2Seq and Vid2Seq (V)} 
We notice that the original Vid2Seq paper~\citep{yang2023vid2seq} also reports the performance of Vid2Seq on the VPC task, but we have confirmed that the results are obtained by removing timestamp outputs from dense captioning outputs and they are inferior to the results we get by fine-tuning the Vid2Seq weight specifically on the VPC task where the inputs are $V$ and $A$.
Therefore, we report our fine-tuning results as the performance of Vid2Seq in the main text.
Moreover, to get a competitive baseline in the video-only setting, 
we fine-tune the Vid2Seq pretrained weight in this setting and report the results as the performance of Vid2Seq (V) in the main text.
The training schemes for Vid2Seq and Vid2Seq (V) follow the setup stated in \S~\ref{sunsec:exp_setup}.

\paragraph{Our MVPC and MR-VPC}
During inference, we apply a length penalty of 1.0 for YouCook2 and ActivityNet Captions, and a length penalty of 0.6 for Charades. 
In the \emph{DistillAM} strategy, when utilizing the MVPC model to generate training data for training the MR-VPC model, we keep the same inference hyperparameters.
Notably, we notice that unfreezing top CLIP layers has minimal impact on the performance of MVPC and VidSeq in our preliminary experiments, but the choice significantly boosts the performance of MR-VPC. 
Thus, we unfreeze the last six CLIP layers in the video encoder when training MR-VPC models.
In this situation, the total trainable parameters are 390M.~\looseness=-1

\section{Details of Model-Based Metrics} \label{app:metrics}
We use the following model-based automatic evaluation metrics:

\begin{itemize}
    \item \textbf{Perplexity (PPL)}: To assess the fluency of the generated paragraph-level captions, we adopt the perplexity score produced by a pretrained language model gpt2-large~\citep{gpt2} (774M parameters).
    \item  \textbf{BERTScore }~\citep{zhangbertscore} and \textbf{BARTScore}~\citep{yuan2021bartscore} are two text generation metrics based on the similarities of BERT~\citep{devlin-etal-2019-bert} embeddings and the generation probabilities of the BART~\citep{lewis2020bart} model, respectively. We use them for evaluating the consistency between generated captions and reference captions.
    Specifically, for BERTScore, we use the F1 score given by the roberta-large~\citep{roberta} pretrained model (335M parameters); for BART score, we use the facebook/bart-large-cnn model (406M parameters) trained on ParaBank2~\citep{hu2019large}.~\footnote{Available at  \url{https://github.com/neulab/BARTScore}.}~\looseness=-1
    \item  \textbf{EMScore}~\citep{shi2022emscore} is an automatic video captioning metric derived by matching the video frame embeddings and the text token embeddings produced by the CLIP~\citep{radford2021learning} model. 
    Besides the reference-free version EMScore (EMS for short), we also report the reference-based version EMS$_{\text{ref}}$ additionally considering the similarity of the prediction and the reference annotation.  
    Concretely, we use the clip-vit-base-patch32 pretrained model (151M parameters) following \citet{shi2022emscore}.\looseness=-1
\end{itemize}

\section{Effect of Drop Rates $\mathbf{p_A}$ and $\mathbf{p_E}$} \label{app:drop_rate}
Recall that $p_A$ and $p_E$ are the probabilities to be nullified for the ASR modality $A$ and the event boundary modality $E$ in our \emph{DropAM} strategy in \S~\ref{method:drop}.
We enumerate the values of these two hyperparameters (called drop rates) in \emph{DropAM} and report the CIDEr results on the validation set of YouCook2 in Table~\ref{tab:drop_rates}.
We observe that large drop rates hamper performance in the modality-complete setting and small drop rates result in poor performance in the modality-incomplete setting.
Generally, setting $p_A$ and $p_E$ around 0.5 strikes the balance relatively well and performs the best in terms of the average performance with different available modalities.
Moreover, we have made similar observations on ActivityNet Captions in preliminary explorations.
Therefore, we use $p_A = p_E =$ 0.5 in our main experiments.

\begin{table}[t] \small
\centering
\resizebox{0.48\textwidth}{!}{
\begin{tabular}{@{}cc|cccc|c@{}}
\toprule
\multirow{2.5}{*}{$\mathbf{p_A}$} & \multirow{2.5}{*}{$\mathbf{p_E}$} &  \multicolumn{4}{c|}{\textbf{Test Modalities}} &   \multirow{2.5}{*}{\textbf{Avg.}}    \\
 \cmidrule(lr){3-6}
   &  & \textbf{V+E+A} & \textbf{V+E} & \textbf{V+A} & \textbf{V}  &\\ \midrule
     0.1            &     0.1            &    23.15            &    13.50          &    21.80     &  10.76   &    17.30     \\
  0.3            &     0.3            &     23.04           &    15.76          &     22.52        &     15.39    & 19.18  \\
      0.5           &   0.5              &  22.67             &    16.94         & 22.54         &       16.53 & \textbf{19.67}  \\
   0.7  &      0.7               &    22.24         &   17.10       &   22.30      &   17.03   & \textbf{19.67}    \\
     0.9            &     0.9           & 19.86               &     17.52         &     19.84        &      17.57 &  18.70   \\
   \bottomrule
\end{tabular}}
\caption{The effect of the choice of drop rate $p_A$ and $p_E$ with different available modalities at test time on the YouCook2 dataset. Only the \emph{DropAM} strategy is applied and METEOR metrics are reported.}
\label{tab:drop_rates}
\vspace{-0.5cm}
\end{table}

\section{Comparison with \citet{yang2023vidchapters}} \label{app:vidchap}
Concurrent to our work, \citet{yang2023vidchapters} extend Vid2Seq to incorporate both $A$ and $E$ for VPC (called \textit{``video chapter generation given ground-truth boundaries''} in their paper).
Specifically, they trim long videos into short clips given the ground-truth event boundaries $E$, train Vid2Seq on the short clips for sentence-level captioning, and concatenate the predictions on each clip to form paragraph-level captions.
The proposal by \citet{yang2023vidchapters} (named as ``Vid2Seq-Concat'' by us) has two weaknesses: (1) Vid2Seq-Concat simply divides the VPC task into video captioning on short clips and fails to model the inter-event dependence in each long video; (2) the video and ASR input of Vid2Seq-Concat is determined by the given event boundaries, which makes the system vulnerable when the event boundaries are noisy or absent.
In comparison, our MVPC and MR-VPC schemes model all input modalities in an end-to-end manner, bringing two key advantages: (1) effective modeling of inter-event dependence in long videos; (2) no information loss in $A$ and $V$ when $E$ is noisy.
The experimental results on YouCook2 in Table~\ref{tab:vid_chap_comp} empirically validate the advantage of our proposals over Vid2Seq-Concat~\citep{yang2023vidchapters}\footnote{We uniformly sample 15 frames for each clip in our implementation of Vid2Seq-Concat to keep the total input frames of each video close to 100.}.

\begin{table}[t] \small
\centering
\resizebox{0.48\textwidth}{!}{
\begin{tabular}{@{}l|cccc|c@{}}
\toprule
\multirow{2.5}{*}{\textbf{Model}}  &  \multicolumn{4}{c|}{\textbf{Test Modalities}} &   \multirow{2.5}{*}{\textbf{Avg.}}    \\
 \cmidrule(lr){2-5}
   & \textbf{V+E+A} & \textbf{V+E} & \textbf{V+A} & \textbf{V}  &\\ \midrule
     VidSeq-Concat                     &   22.35            &        14.08    & 21.92      &     12.12  &   17.62     \\
      MR-VPC (Ours)                 &   \textbf{22.83}         &      \textbf{16.97}     &    \textbf{22.59}   &    \textbf{16.86}   &  \textbf{19.81}  \\
   \bottomrule
\end{tabular}}
\caption{Comparison with Vid2Seq-Concat~\citep{yang2023vidchapters} on YouCook2. METEOR metrics are reported. When testing Vid2Seq-Concat without $E$, we trim the video into seven consecutive clips of the same length (seven is the average number of events in YouCook2).}
\label{tab:vid_chap_comp}
\vspace{-0.5cm}
\end{table}

\begin{figure*}[t]
\centering
\includegraphics[width=0.95\textwidth]{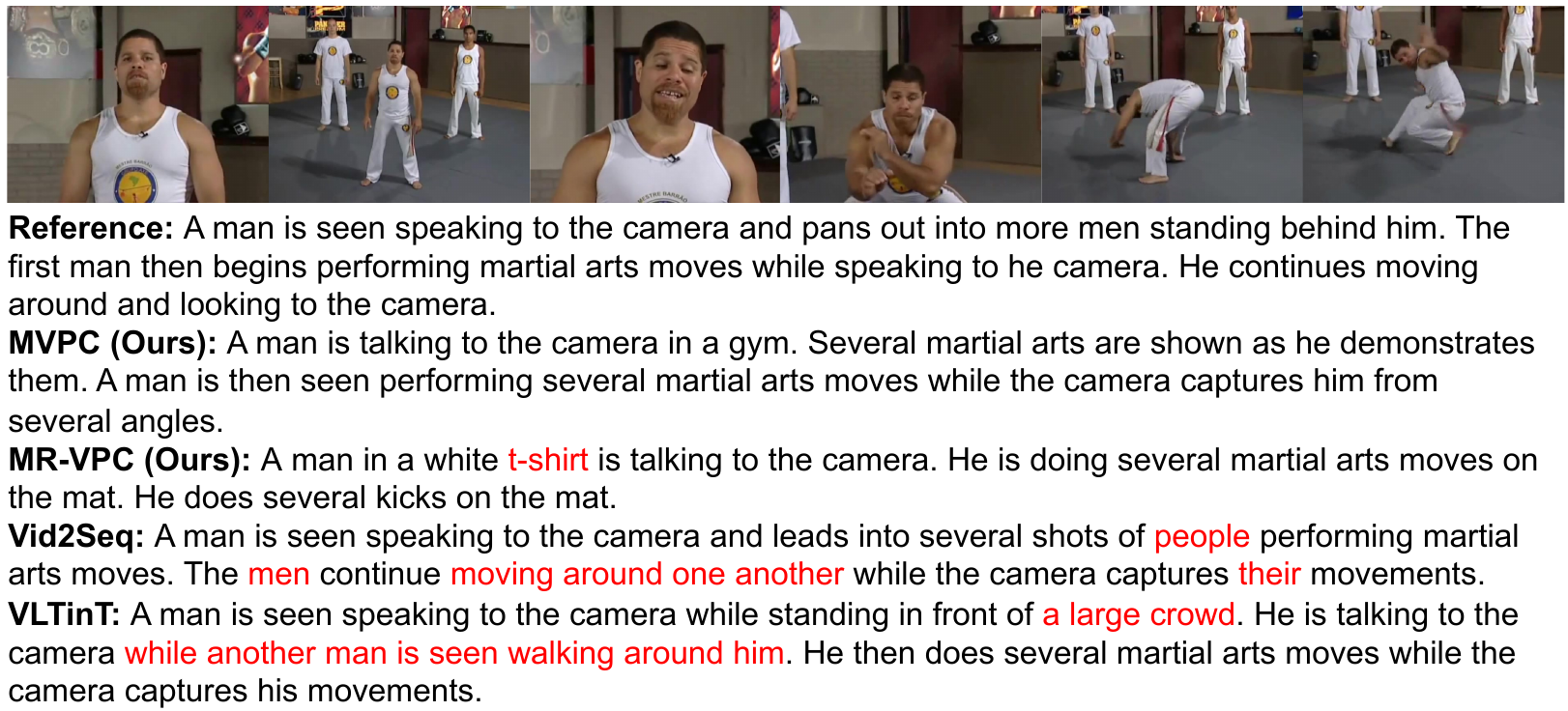}
\caption{The captions produced by our models and baselines in the modality-complete setting on an ActivityNet Captions test sample (id: ``bXdq2zI1Ms0''). The \textcolor{red}{wrongly predicted concepts} are highlighted in \textcolor{red}{red} by the author.}
\label{fig:demo1}
\end{figure*} 

\begin{table*}[t]
\centering
\resizebox{0.95\textwidth}{!}{
\begin{tabular}{@{}l|l|l@{}}
\toprule
\textbf{Model Predictions} & \textbf{Modality-Complete Setting}                                                                                                                                    & \textbf{Video-Only Setting}                                                                                                                              \\ \midrule
Reference                 & \multicolumn{2}{l}{pick the ends off the verdalago. combine lemon juice sumac garlic salt and oil in a bowl. chop lettuce and place it in a bowl. $\cdots$}                                                                                                                                                     \\
\midrule
Vid2Seq                    & \begin{tabular}[c]{@{}l@{}}wash the leaves of verdolago. add lemon juice sumac crushed \\ garlic salt and olive oil to a bowl and mix. $\cdots$\end{tabular} & { \begin{tabular}[c]{@{}l@{}}\textcolor{red}{um, i'ma add some sea salt to the bowl. add some black} \\ \textcolor{red}{ pepper and mix it well. $\cdots$} \end{tabular}} \\
\midrule
MVPC                       & \begin{tabular}[c]{@{}l@{}}wash the pita bread slices. mix lemon juice sumac garlic salt and \\ olive oil in a bowl. $\cdots$\end{tabular}                   & \textcolor{red}{tv.sv.svs.svv.svv on svvvm.svvm on svhvm on the svvm.}                                                                      \\ \midrule
MR-VPC                     & \begin{tabular}[c]{@{}l@{}}wash the romaine lettuce leaves. add lemon juice sumac crushed \\ garlic salt and olive oil to a bowl. $\cdots$ \end{tabular}      & \begin{tabular}[c]{@{}l@{}}wash the romaine lettuce leaves. add lemon juice sumac \\ crushed garlic salt and olive oil to a bowl. $\cdots$\end{tabular}   \\  \bottomrule
\end{tabular}}
\caption{The predictions given by the models on a YouCook2 instance (id: ``xHr8X2Wpmno'')  in the modality-complete setting (the second column) and the video-only setting (the third column). We only show the first two sentences of the predictions due to the limit of space and the \textcolor{red}{degenerated predictions } are highlighted in \textcolor{red}{red}. }
\label{tab:pred_case}
\end{table*}

\section{Noise Besides Missing Modality} \label{app:noise}
In \S~\ref{subsec:other_noise}, we discuss five types of noise in auxiliary modalities. Here are the details of them:
\begin{itemize}
    \item \textbf{Low-quality ASR}: In real-world scenarios, ASR systems may have hardware limitations, resulting in inferior ASR data compared to the ASR texts used during training generated by state-of-the-art ASR models. 
    To simulate this situation, we replace the Whisper small.en model (244M parameters) with the tiny.en model (39M) and reduce the inference beam size from 5 to 1.
    \item  \textbf{ASR Sentence Deletion}: To simulate the corruption of ASR data, we randomly delete 50\% of all sentences in each test instance.
    \item \textbf{Event Deletion}: In order to simulate the corruption of event boundary data, we randomly delete 50\% of the events in the event boundary data of each test instance.
    \item \textbf{Boundary Perturbation}: To introduce perturbations to the event boundaries, we add random uniform noise ranging from -5 to +5 units (percentage points) to each timestamp in the event boundaries of each instance.
    \item \textbf{Generated Boundary}: Considering that event boundaries predicted by models are more realistic noisy inputs than perturbed ground-truth boundaries, we leverage the PDVC~\citep{wang2021end} dense captioning model to generate event boundaries.
\end{itemize}

\begin{figure}[t]
\centering
\includegraphics[width=0.48\textwidth]{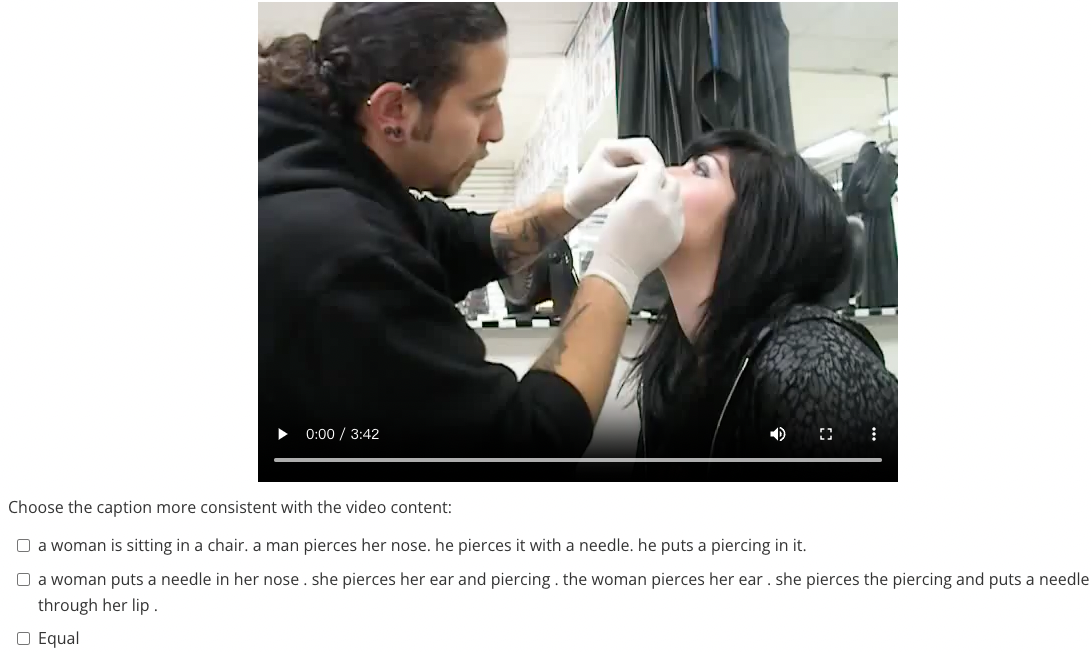}
\caption{The human annotation interface.}
\label{fig:UI}
\end{figure} 

\section{Software and Hardware Requirements}
We implement our code based on the PyTorch \citep{torch} and HuggingFace Transformers \citep{wolf2020transformers} Python libraries. 
All experiments in this paper are conducted on a server with 8 NVIDIA A40 GPUs (48 GB memory per GPU).

\section{Qualitive Example} \label{app:qual}
We present a qualitative case study in Figure~\ref{fig:demo1} to highlight the strengths of our MVPC and MR-VPC models in the modality-complete setting.
As shown, VidSeq and VLTinT baselines tend to produce hallucinations and predict concepts inconsistent with the video content.
For example, although there is only one man moving and performing martial arts in the video, Vid2Seq predicts ``\emph{The men continue moving around one another}'' and VLTinT generates ``\emph{another man is seen walking around him}''.
In contrast, our MVPC and MR-VPC models show almost no hallucinations. 
The generated captions are more accurate and closely aligned with the content of the video.

\section{Details of Human Evaluation} \label{app:human}

Three voluntary annotators, who are graduate students fluent in English, are asked to choose a caption that they deem more coherent with the video content from a pair of model predictions or choose the ``equal'' option if they consider the two predictions to be equally good in terms of coherence.
The data collection protocol is approved by an internal ethics review.
We depict the layout of the annotation webpage in Figure~\ref{fig:UI}.~\looseness=-1

\end{document}